\documentclass{article}
\usepackage{spconf,amsmath,graphicx}
\usepackage{mathtools}
\usepackage{amssymb}
\usepackage{amsthm}
\usepackage{thmtools}
\usepackage{times}
\usepackage{epsfig}
\usepackage{graphicx}
\usepackage{amsmath}
\usepackage{amssymb}
\usepackage{booktabs}
\usepackage{hyperref}
\usepackage{caption}
\usepackage{color}
\usepackage{multirow}
\usepackage{graphicx}
\usepackage{multirow}
\usepackage{graphicx,multirow}
\usepackage{wrapfig}
\usepackage{subcaption}
\usepackage{enumitem}
\setlist{nosep, leftmargin=14pt}

\usepackage{mwe} 
\usepackage{bm}
\usepackage{dsfont}

\title{Open-Set Semi-Supervised Learning for Long-Tailed Medical Datasets}
\name{Daniya~Najiha~A.~Kareem$^1$, Jean~Lahoud$^1$, Mustansar~Fiaz$^2$, Amandeep~Kumar$^3$, and Hisham~Cholakkal$^1$ 
}
\address{$^1$Mohamed bin Zayed University of AI, UAE \hspace{0.5mm} $^2$IBM Research, UAE \hspace{1.5mm} $^3$Johns Hopkins University, US}
\begin{document}
%
\maketitle
\begin{abstract}
Many practical medical imaging scenarios include categories that are under-represented but still crucial.
The relevance of image recognition models to real-world applications lies in their ability to generalize to these rare classes as well as unseen classes. 
Real-world generalization requires taking into account the various complexities that can be encountered in the real-world. First, training data is highly imbalanced, which may lead to model exhibiting bias toward the more frequently represented classes. Moreover, real-world data may contain unseen classes that need to be identified, and model performance is affected by the data scarcity. 
While medical image recognition has been extensively addressed in the literature, current methods do not take into account all the intricacies in the real-world scenarios.
To this end, we propose an open-set learning method for highly imbalanced medical datasets using a semi-supervised approach. 
Understanding the adverse impact of long-tail distribution at the inherent model characteristics, we implement a regularization strategy at the feature level complemented by a classifier normalization technique.	
We conduct extensive experiments on the publicly available datasets, ISIC2018, ISIC2019, and TissueMNIST with various numbers of labelled samples. 
Our analysis shows that addressing the impact of long-tail data in classification significantly improves the overall performance of the network in terms of closed-set and open-set accuracies on all datasets.
Our code and trained models will be made publicly available at \url{https://github.com/Daniyanaj/OpenLTR}.

\end{abstract}

\section{Introduction}
\label{sec:intro}

Medical image recognition includes classifying images, such as pathological images, X-rays, MRI scans, and CT scans into different classes to aid in diagnosis, disease monitoring, and treatment planning \cite{vmixer,dwinformer}.
In the practical scenarios, the classification task faces numerous challenges. 
Often, labelled data for categories of interest is limited, where high-quality annotations are scarce due to privacy, ethical concerns, and the requirement for expert annotations.
In addition, medical images are complex due to the fine-grained nature which often contain subtle features, thus requiring advanced analysis techniques and high-resolution imaging to accurately interpret and diagnose conditions. 
Imbalanced class distribution poses another hurdle, as disease-positive cases are relatively scarce within the broader population, leading to a large variation in the occurrence rate of different diseases. 



To overcome the challenge of limited annotations, semi-supervised learning (SSL) has been introduced as a promising approach \cite{mao2022pseudo, che2017boosting}, \cite{haghighi2022dira},  and \cite{liu2021self, unnikrishnan2021semi}.
While SSL approaches offer potential solutions to the limited data availability, 
they do not address other challenges, such as the class imbalance and the need to detect novel classes not seen in the training phase. Effectively managing this class imbalance is of paramount importance to prevent false negatives in the minority class (i.e., the disease-positive cases), as overlooking them could lead to potentially fatal outcomes.
In many real-world medical scenarios, it's often impractical to manually acquire such an extensive dataset with equal distribution among all disease classes. The imbalance in dataset affects the natural equilibrium state of the classifier feature representations. 
When training with imbalanced data, it is important to carefully balance the distributions at both the feature and classifier levels to ensure effective learning \cite{kothapalli2022neural, alshammari2022long, singh2020investigating}. 


A more complex scenario emerges, wherein unlabelled data may contain outliers representing unseen classes not present in the labelled dataset. 
%
To this end, we introduce a novel recognition system that accurately classifies both prevalent and less common classes, generalizes from limited examples of known instances, and correctly identifies novel instances that have not been encountered before. We test our method in a practical real-world scenario, where data annotation is limited and unbalanced, and novel classes might appear at inference time.
In summary, this paper has the following contributions:
\begin{itemize}

 \item An open-set SSL framework is adopted for medical image classification that can classify known categories and can distinguish between known and unknown classes, in a unified paradigm. 
 \item  To rule out the adverse effect of data imbalance on classification, we implement a regularization strategy at the feature level in addition to weight normalization at the classifier layer. 

\item Extensive experiments are conducted on three distinct datasets to showcase the efficacy of our techniques aimed at tackling class imbalance issues in both supervised learning and SSL contexts.

\end{itemize}

\section{Methodology}
\label{sec:method} 

\begin{figure*}[t]
\begin{center}
\includegraphics[width=1.0\linewidth,trim={0 4mm 0 4mm},clip]{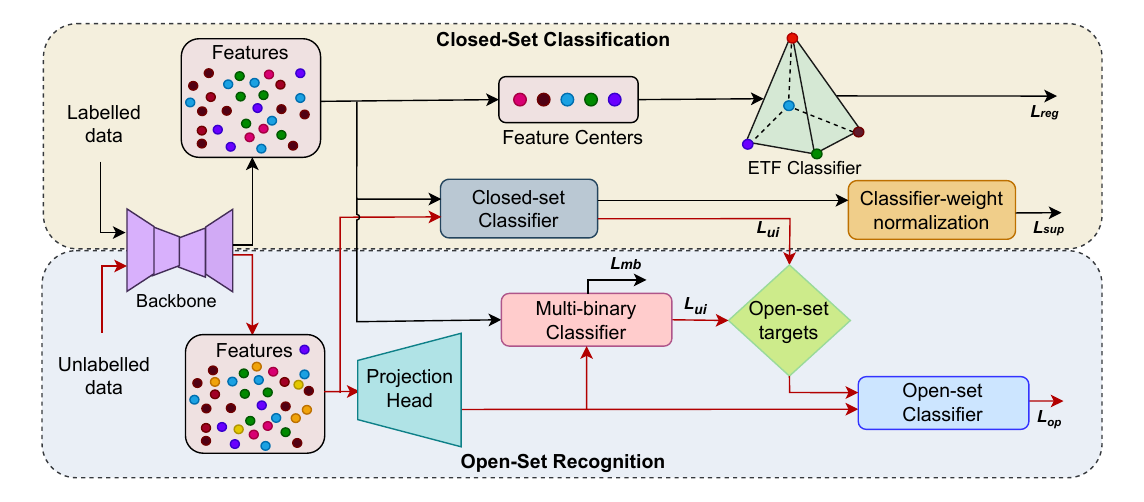}
\end{center}
\vspace{-0.4cm}   
   \caption{\textbf{Overall Architecture}: The framework is organized into two branches, namely, a closed-set classification branch and an open-set recognition branch. The black and red arrows correspond to the flow of labelled data and unlabelled data respectively. In the closed-set classification branch, the backbone features of the labelled batch are passed to a closed-set classifier and to a multi-binary classifier, which are optimised by the cross-entropy and multi-binary losses respectively. In parallel, for feature regularization, another classifier initialised as a random simplex equiangular tight frame (ETF) is applied on the feature centers of the labelled batch and optimised using the regularization loss. Further, a classifier weight normalization technique is employed on the closed-set classifier weights to overcome the adverse effect of data imbalance on training. In the open-set recognition branch, features from the unlabelled batch are projected into a low dimensional feature embedding on which a set of multi-binary classifiers are employed. The multi-binary classifier along with closed-set classifier produces open-set targets. These open-set targets optimize the closed-set and open-set classifiers to achieve joint inliers and outliers utilization.}
\vspace{-0.5cm}   
\label{fig:arch}
\end{figure*}

We propose an integrated framework to classify known categories in the long tail medical data from a few samples and identify the unknown classes within the dataset. The input images, denoted by  $\mathcal{Y}\in \mathcal{R}^{ H \times W \times D}$, are processed to form labelled batch $\mathcal{I_C}$ and unlabelled batch $\mathcal{I_U}$. 
Our network is organized into two main branches, a closed-set classification branch and an open-set recognition branch, as shown in Figure \ref{fig:arch}.
Both the closed-set and open-set classification branches are trained simultaneously to classify the known classes from a few samples, as well as to identify unknown classes.

\vspace{-0.3cm}
\subsection{Closed-Set Classification Branch}

In the closed-set classification branch, features extracted from the labelled batch with seen-class samples are fed to a multi-class classifier, which is then optimized with the cross-entropy loss $\text{H}(\cdot,\cdot)$:
\vspace{-0.4cm}
\begin{equation}
\label{eq:loss-s}
    \mathcal{L}_{sup}(\mathcal{I_C})=\frac{1}{M}\sum_{i=1}^M \text{H}(\bm{z}_i,y_i).
\end{equation} 
\noindent where $\bm{M}$ is the batch size, $z_i$ is the classifier output, and ${y}_i$ is the corresponding ground-truth label. 


\noindent\textbf{Feature-Center Regularization:}
In a training setup with class-balanced data, during the final stage of training, the features of the final layer converge towards the mean values within their respective classes \cite{papyan2020prevalence}. These within-class means, combined with the classifier weights, will then collapse to the vertices of a simplex that forms an equiangular tight frame (ETF).


 For a  dataset with L classes, let $v_i \in  { R}^d$ be the d-dimensional feature of the final classifier layer. Simplex equiangular tight frame (ETF) can be defined as a collection of vectors namely, ${n}_l\in\mathbb{R}^d$, $l=1,2, ..., L$, $d\ge L$, which upon satisfying the below condition is called  a simplex ETF. If  ${N}=[{n}_1, ...,{n}_K]\in\mathbb{R}^{d\times L}$, ${U}\in\mathbb{R}^{d\times L}$ allows a rotation and satisfies ${U}^\top{U}={I}_L$, ${I}_L$ is the identity matrix, and ${1}_L$ is a vector with only ones,

	\begin{equation}\label{ETF_M}
		{N}=\sqrt{\frac{L}{L-1}}{U}\left({I}_L-\frac{1}{L}{1}_L{1}_L^\top\right),
	\end{equation}
	
All vectors in a simplex ETF have the same pair-wise angle and an equal $\ell_2$ norm with 
\begin{equation}\label{mimj}
	{n}_i^\top{n}_j=\frac{L}{L-1}\delta_{i,j}-\frac{1}{L-1}, \forall i, j \in \{1, ..., L\},
\end{equation}
where $\delta_{i,j}$ equals to $1$ when $i=j$ and $0$ otherwise. The pair-wise angle $-\frac{1}{K-1}$ is the maximal equiangular separation of $K$ vectors in $\mathbb{R}^d, d\ge K-1$ \cite{nc}. To preserve this maximally separated and equiangular structure of the feature centers , we employ classifier regularization strategy that extracts
the feature centers $\bar{v}_k $ of each class. These feature centers are then concatenated to a single vector labels $\bar{{V}}$ and passed on to the classifier layer $B^*.$ This classifier layer is fixed as a simplex
ETF to regularize the feature centers. 
This fixed classifier helps in the alignment of feature centers to achieve the equiangular separation and enhances the discriminative ability among the tail classes.  The classifier output $\bar{{z}}$ and the label centers $\bar{{y}}$  are then used to measure the degree of feature center collapse. 
Then, the $\mathcal{L}_{\rm reg}$ loss can be written as follows:
\vspace{-0.3cm}
\begin{equation}
       \mathcal{L}_{\rm reg}(\bar{{z}}, \bar{{y}})=\frac{1}{M}\sum_{i=1}^M \text{H}(\bar{{z}},\bar{{y}}).
\end{equation}


\noindent\textbf{Classifier Weight Normalization:}
In addition to the feature regularization, we also apply regularization to the closed-set classifier weights to constrain the norms of weight vectors in the neural network.
This helps prevent overfitting and enhances generalization by keeping the weights from growing excessively large.
The regularization limits the maximum weight values in the network by projecting the weights onto a ball of a specified radius if weight norm values exceed this radius \cite{alshammari2022long}. It constrains the weight vector ${w}$ such that:
\vspace{-0.3cm}
\[
\text{if } \|{w}\|_2 > a, \text{ then } {w} \leftarrow {w} \cdot \frac{a}{\|{w}\|_2}
.\]

\noindent where $\|{w}\|_2$ is the Euclidean norm of the weight vector ${w}$, and $a$ is the maximum allowed norm.
\vspace{-0.3cm}
 \subsection{Semi-Supervised Branch}
 To get a robust prediction for the unseen class samples, a multi-binary classifier is also implemented on the labelled batch $\mathcal{I_L}$. The multi-binary classifier can be seen as a combination of $K$ binary classifiers. To optimize the multi-binary classifier, features are projected into another feature space with the projection head $f(\cdot)$.
The classifier performs a one-vs-rest classification for each of the known classes to output the class-wise likelihood of outliers or inliers with respect to the $k$-th seen class. The hard-negative sampling strategy \cite{saito2021ovanet} is adopted to optimize the multi-binary classifier with the labelled samples. Let (${o}_{i,k}, \bar{o}_{i,k} $) be the binary classifier output for the $k$-th seen class, then the multi-binary loss is given by:
\vspace{-0.3cm}
\begin{equation}
\label{eq:loss-mb}
    \mathcal{L}_{mb}(\mathcal{I_C})=\frac{1}{M}\sum_{i=1}^M \left( -\log(o_{i,y_i})-\min_{k\neq y_i}\log(\bar{o}_{i,k})\right).
\end{equation}  

The predictions of closed-set and multi-binary classifiers are fused to get the open-set targets. Specifically, for each unlabelled sample $\bm{u}_i$, the seen-class probability distribution is predicted by the closed-set classifier ($\widetilde{z}_{i,k}$) and multi-binary classifier ($o^w_{i,k}$).   
These distinct and complementary predictions determine the likelihood that $\bm{u}_i$ belongs to one of the known classes or is an outlier. Therefore,
\vspace{-0.2cm}
\begin{equation}
\label{eq:target-q}
    \widetilde{r}_{i,k} = 
        \begin{cases}
         \widetilde{z}_{i,k} \cdot o^w_{i,k}  & \quad \text{if\quad} 1\leq k \leq K;
         \\[10pt] 
         \sum_{j=1}^{K} \widetilde{z}_{i,j} \cdot \bar{o}^w_{i,j} & \quad \text{if\quad} k = K + 1. 
        \end{cases}
\end{equation}

In this manner, without explicitly distinguishing between inliers and outliers, open-set targets can be generated.
These targets can then be used to train the open-set classifier.
\vspace{-0.2cm}
\begin{equation}
\label{eq:loss-op}
    \mathcal{L}_{o}(\mathcal{I_U})=\frac{1}{\mu M}\sum_{i=1}^{\mu M} \mathds{1}(\max_k(\widetilde{r}_{i,k})>\tau_r) \cdot \text{H}(\bm{\widetilde{r}}_i,\bm{r}^s_i)
\end{equation} 
where $\bm{r}_i^s$ are the predictions on the strongly augmented samples and $\tau_r$ is the confidence threshold. 
The two classifiers are further optimised with the help of a double filtering strategy for closed-set classifier to pick best class pseudo-labels of inliers from known classes.

\vspace{-0.3cm}
\begin{equation}
\label{eq:loss-u}
    \mathcal{L}_{ui}(\mathcal{I_U})=\frac{1}{\mu M}\sum_{i=1}^{\mu M} \mathcal{F}(\bm{u}_i)\cdot \text{H}(\widetilde{\bm{z}}_i,\bm{z}^s_i).
\end{equation} 
where $\mathcal{F}(\cdot)$ is the filtering function based on a confidence threshold $\tau_p$. $\mathcal{L}_{ui}$ helps to exclude highly probable outliers and incorrect pseudo-labels of inliers, which eventually helps in identifying the known and unknown category samples.

As seen in equation \ref{eq:loss-op},  $\mathcal{L}_{o}$ relies on the open-set targets collectively produced by the closed-set and multi-binary classifiers. Hence, long-tail distribution affects the performance of closed-set classifier which in-turn affects the open-set targets. The feature regularization and weight normalization techniques implemented on the closed-set classification branch helps the classifier to make balanced predictions, consequently improving the quality of open-set targets. 


\begin{table}[t]
\centering
\scalebox{0.75}{
\begin{tabular}{|c|cc|cc|}
\hline
          & \multicolumn{2}{c|}{\textbf{ISIC2018}}     & \multicolumn{2}{c|}{\textbf{ISIC2019}}     \\ \hline
Method    & \multicolumn{1}{c|}{Closed-Set} & Open-Set & \multicolumn{1}{c|}{Closed-Set} & Open-Set \\ \hline
MixMatch \cite{berthelot2019mixmatch}  & \multicolumn{1}{c|}{82.94}      & -        & \multicolumn{1}{c|}{77.17}      &       -   \\ \hline
FlexMatch \cite{zhang2021flexmatch} & \multicolumn{1}{c|}{81.46}      & -        & \multicolumn{1}{c|}{74.62}      &     -     \\ \hline
SimMatch \cite{zheng2022simmatch} & \multicolumn{1}{c|}{82.71}      & -        & \multicolumn{1}{c|}{76.00}      &      -    \\ \hline
CoMatch \cite{li2021comatch}  & \multicolumn{1}{c|}{76.49}      & -        & \multicolumn{1}{c|}{71.63}      &        -  \\ \hline
OpenMatch \cite{saito2021openmatch} & \multicolumn{1}{c|}{79.12}      & 9.94     & \multicolumn{1}{c|}{76.62}      & 11.31    \\ \hline
IoMatch \cite{li2023iomatch}   & \multicolumn{1}{c|}{82.70}      & 36.75    & \multicolumn{1}{c|}{77.22}      & 40.10    \\ \hline
Ours      & \multicolumn{1}{c|}{\textbf{84.08}}      & \textbf{41.16}    & \multicolumn{1}{c|}{\textbf{79.16}}      & \textbf{40.90}    \\ \hline
\end{tabular}
}
\vspace{-0.1cm}
\caption{Performance comparison  on ISIC2018 and ISIC2019 datasets (based on accuracy). The model was trained on 25\% labelled data, using classes (0, 5) in ISIC2018 dataset and classes (0, 6) in ISIC2019 dataset while rest of the classes were not seen during training. \textbf{Closed-set accuracy} refers to  accuracy on classes seen during training while \textbf{open-set accuracy} denotes performance on the open-set data. }
\vspace{-0.6cm}
\label{tab1}
\end{table}

\noindent\textbf{Classifier Weight Balancing:} Weight decay is a prominent form of regularization that applies the penalty of L2 norm on network weights \cite{alshammari2022long}.   It helps in generalizing the model by constraining the growth of weights during training.  Implementing the weight
balancing constrains the weights to a range of values and
helps in balancing the classifier.
\vspace{0.1cm}
\noindent\textbf{Implementation:}
Our method is implemented with PyTorch library and trained on an NVIDIA A100 GPU.
We use ResNet18 \cite{he2016deep} as the backbone and the embedding dimension is set to 128. 
We train the network, for
100 epochs with a batch size of 16 and an initial learning rate of 0.03, with the help of SGD. We conduct experiments with 50\%, 25 \%, and 10 \% labels while training, and the whole dataset without labels is fed to the network as the unlabelled counterpart. The models with the highest accuracy on a
validation set are chosen for testing. Our method is evaluated on three publicly
available skin lesion classification datasets, namely ISIC2018  \cite{isic2018, isic20182}, ISIC2019 \cite{isic2018,  isic20182, isic2019}, and TissueMNIST \cite{tissuemnist}. 

\vspace{-0.3cm}
\section{Results}
\vspace{-0.1cm}

\begin{table}[t]
\centering
\scalebox{0.75}{
\begin{tabular}{|l|c|c|}
\hline
\multicolumn{3}{|c|}{\textbf{ TissueMNIST}}  \\ \hline
\multicolumn{1}{|c|}{Method} & Closed-Set & Open-Set  \\ 
\hline

OpenMatch    { \cite{saito2021openmatch} }             &      68.31              &          14.32          \\ \hline
IoMatch     { \cite{li2023iomatch}     }           & 71.72               & 40.86             \\ \hline
Ours                         & \textbf{72.30}               & \textbf{40.93}             \\ \hline
\end{tabular}
}
\vspace{-0.2cm}
\caption{Performance comparison on the TissueMNIST dataset in terms of accuracy. Models were trained with 25\% labelled data of classes (2,8) while the remaining 2 classes were not seen during training.}
\label{tab5}
\vspace{-0.6cm}
\end{table}
We benchmark our method against the top-performing semi-supervised models in the literature. All the networks were trained on 25\% labelled data with ResNet18 backbone for 100 epochs with similar hyper-parameters.  The performance of our method on the ISIC2018 and ISIC2019 datasets is shown in Table \ref{tab1}. 
Our method has a clear improvement over the state-of-the-art SSL methods in terms of closed-set accuracy  and open-set accuracy. 
Similarly, over TissueMNIST dataset \cite{tissuemnist} where 2 head classes with 13800 and 9800 samples, 2 tail classes with 1466 and 1926 samples, and the rest of the classes contain samples in the range (2500, 6500). Table \ref{tab5} shows that our method outperforms the existing methods.
\vspace{-0.3cm}

\section{Discussion and Ablations}

\vspace{-0.1cm}

\noindent\textbf{Semi-Supervised Learning:} We here study the lower and upper bounds for the semi-supervised learning task.
In our network setup, we train the network only with samples of five classes with the remaining reserved for open-set classification. The lower bound accuracies for this data partition were found as 82.6 (Exp 1) and  80.81 (Exp 2) as shown in Table \ref{tab3}. Experiments 1 and 2 were trained with 25\% of labelled samples with multiclass classifier and multi-binary classifier respectively. As an upper-bound, we train the network with whole training data for the first five classes, employing a multi-class classifier, which showed an accuracy of 86.65 (Exp 5). The open-set accuracies for the above experiments were obtained by applying a hard threshold during the evaluation phase. On ISIC2018, our method yields higher score than the lower bound and achieves a closer value to the upper-bound with minimal percentage of labelled training data.  
\\
\noindent\textbf{Open-Set Learning:}
As presented in Table \ref{tab3}, the lower bounds for open-set recognition task was found as 10.32 \%  ( Exp 1) with the closed-set classifier and 12.1\% (Exp 2) with the multi-binary classifier, both trained with 25\% labelled samples of classes 0 to 5. The maximum achievable accuracy or upper-bound that utilize 100 \% of data with class labels (of all 7 classes) was found to be 43.8 (Exp 4). Apparently, by optimizing the network with the open-set loss, a fair accuracy score is obtained for the open-set recognition task compared to the upper-bound accuracy.

\begin{table}[]
\centering
\scalebox{0.75}{
\begin{tabular}{|c|c|c|c|c|c|c|}
\hline
   & Supervision & classes & labelled \% & Loss & Acc (0,5) & Acc(6,7) \\ \hline
1 & Semi-supervised & 5            & 25                   & CE   & 82.60        & 10.32         \\ \hline
2 & Semi-supervised & 5            & 25                   & BCE  & 80.81          & 12.11          \\ \hline
3 & Semi-supervised  & 7            & 25                   & CE   & 81.69          & 32.09         \\ \hline
4 & Fully-supervised& 7            & 100                  & CE   & 87.90           & 43.80          \\ \hline
5& Fully-supervised  & 5            & 100                  & CE   & 86.65          & 15.48         \\ \hline

\end{tabular}}
\vspace{-0.2cm}
\caption{Analysis of the upper and lower-bounds for semi-supervised learning and open-set recognition. Acc (0,5) denotes the accuracy on classes 0 to 5 and Acc (6,7) denotes accuracy on classes 6 and 7. }
\label{tab3}
\vspace{-0.25cm}
\end{table}
\begin{table}[]
\centering
\scalebox{0.75}{
\begin{tabular}{|c|c|c|}
\hline
  & Technique               & Closed-set Acc \\ \hline
1 & Semi-supervised setting with CE loss & 82.73        \\ \hline
2 & + Weight Decay          &         82.80       \\ \hline
3 & + Feature Regularization  &   83.51   \\ \hline
4 & +    Classifier Weight Normalisation  & \textbf{84.08}         \\ \hline

\end{tabular}}
\vspace{-0.2cm}
\caption{Ablation study on ISIC2018 dataset in terms of accuracy (\%) measure on the classes seen during training.}
\vspace{-0.6cm}
\label{tab4}
\end{table}
\noindent\textbf{Long-Tail Learning:}
A number of methods were proposed to tackle the long-tail classification task, most of which employs loss re-weighting schemes to balance the learning process. In contrast, we employ strategies that do not require class distribution information in the training phase. Table \ref{tab4} presents the contribution of individual techniques to the overall performance of our method. All the experiments were trained with 25\% of labelled data. 

\noindent\textbf{Feature Regularization and Classifier Normalization:}
It is evident that addition of the feature regularization strategy to preserve the inherent properties of the final layer classifier impacts significantly in the model performance. In addition, it helps in producing better open-set targets for the open-set recognition task, consequently improving the open-set classification accuracy. It can be seen from Table \ref{tab1} that, feature regularization at the supervised branch helps in improving the closed-set accuracy by 1.3 \% in ISIC2018 and by 1.9 \% in ISIC2019 datasets. By employing an additional ETF structured classifier in the semi-supervised branch, the tendency of the classifier to bias towards the majority class is prevented, hence improving the classification accuracy for the minor classes.
Classifier weight normalization helped to align classifier weights on a ball-surface and to enhance the growth of small weights within the ball. Furthermore, it can be observed from Table \ref{tab4} that classifier normalization has helped in improving the classification performance of minor classes. 

\vspace{-0.4cm}
\section{Conclusion}
  \vspace{-0.1cm}

\vspace{-0.1cm}
We propose a novel open-set framework that addresses long-tail classification in medical images with few-shot learning. To alleviate the effect of class imbalance, we employ feature regularization and classifier weight normalization. It helps in preserving the innate properties of the classifier exhibited in a balanced data classification task, hence aiding to the performance improvement in rare and unseen classes. Our method shows improved results in three publicly available long-tail medical datasets.

\noindent\textbf{Compliance with Ethical Standards:} This research study was conducted  using human subject data made available in open access by the  ISIC Skin Image Analysis Workshop and Challenge \texttt{@} MICCAI 2018 \cite{isic2018, isic20182},  ISIC Skin Image Analysis Workshop \texttt{@} CVPR 2019 \cite{isic2018,isic20182,isic2019} and TissueMNIST dataset which is an open-source dataset released by \cite{tissuemnist}. For all datasets, ethical approval was not required as confirmed by the license attached with the open access data. \\

\noindent\textbf{Acknowledgement:}
This work is partially supported by the Google Research Award to Hisham Cholakkal, MBZUAI-WIS Joint Program for AI Research (Project grant number- WIS P008), and Meta Llama Impact Innovation Award 2024.

\bibliographystyle{IEEEbib}
\bibliography{refs}

\end{document}